\documentclass[letterpaper]{article} % DO NOT CHANGE THIS
\usepackage{aaai23}  % DO NOT CHANGE THIS
\usepackage{times}  % DO NOT CHANGE THIS
\usepackage{helvet}  % DO NOT CHANGE THIS
\usepackage{courier}  % DO NOT CHANGE THIS
\usepackage[hyphens]{url}  % DO NOT CHANGE THIS
\usepackage{graphicx} % DO NOT CHANGE THIS
\usepackage{natbib}  % DO NOT CHANGE THIS AND DO NOT ADD ANY OPTIONS TO IT
\usepackage{caption} % DO NOT CHANGE THIS AND DO NOT ADD ANY OPTIONS TO IT
\usepackage{algorithm}
\usepackage{algorithmic}

\usepackage{newfloat}
\usepackage{listings}
\DeclareCaptionStyle{ruled}{labelfont=normalfont,labelsep=colon,strut=off} % DO NOT CHANGE THIS
\floatstyle{ruled}
\newfloat{listing}{tb}{lst}{}
\floatname{listing}{Listing}
\usepackage{amsmath}
\usepackage{subfig}
\title{ADEPT: A DEbiasing PrompT Framework}
\author{
    Ke Yang\textsuperscript{\rm 1}, 
    Charles Yu\textsuperscript{\rm 2}, 
    Yi R. Fung\textsuperscript{\rm 2}, 
    Manling Li\textsuperscript{\rm 2}, 
    Heng Ji\textsuperscript{\rm 2}
}
\affiliations{
    \textsuperscript{\rm 1}Tsinghua University\\
    \textsuperscript{\rm 2}University of Illinois Urbana-Champaign\\
    yang-k19@mails.tsinghua.edu.cn\\
    \{ctyu2,yifung2,manling2,hengji\}@illinois.edu\\
}

\usepackage{bibentry}
\begin{document}

\maketitle

\begin{abstract}
Several exisiting approaches have proven that finetuning is an applicable approach for debiasing contextualized word embeddings. Similarly, discrete prompts with semantic meanings have shown to be effective in debiasing tasks. With unfixed mathematical representation at the token level, continuous prompts usually surpass discrete ones at providing a pre-trained language model (PLM) with additional task-specific information. Despite this, relatively few efforts have been made to debias PLMs by prompt tuning with continuous prompts compared to its discrete counterpart.
%\heng{what is the problem on using discrete prompts that continuous prompts can solve?}
% Furthermore, a major problem for debiasing PLMs is the need to not only decrease the bias in the PLM but also to ensure that the PLM does not lose its representation ability. \heng{again why discrete prompt will make it lose representation ability?} 
Furthermore, for most debiasing methods that alter a PLM's original parameters, a major problem is the need to not only decrease the bias in the PLM, but also ensure that the PLM does not lose its representation ability. Finetuning methods typically have a hard time maintaining this balance, as they tend to aggressively remove meanings of attribute words (like the words developing our concepts of ``male" and ``female" for gender), 
% \heng{define what is attribute word?} 
which also leads to an unstable and unpredictable training process. In this paper, we propose \textbf{ADEPT}, a method to debias PLMs using prompt tuning while maintaining the delicate balance between removing biases and ensuring representation ability\footnote{The code and data are publicly available at \url{https://github.com/EmpathYang/ADEPT}.}. To achieve this, we propose a new training criterion inspired by manifold learning and equip it with an explicit debiasing term to optimize prompt tuning. In addition, we conduct several experiments with regard to the reliability, quality, and quantity of a previously proposed attribute training corpus in order to obtain a clearer prototype of a certain attribute, which indicates the attribute's position and relative distances to other words 
% \heng{unclear what 'clearer prototype of certain attribute' means} 
on the manifold. We evaluate \textbf{ADEPT} on several widely acknowledged debiasing benchmarks and downstream tasks, and find that it achieves competitive results while maintaining (and in some cases even improving) the PLM's representation ability. We further visualize words' correlation before and after debiasing a PLM, and give some possible explanations for the visible effects.
\end{abstract}

\section{Introduction}
\begin{figure*}[t]
    \centering
    \subfloat[While debiasing, ADEPT only trains the prompt parameters and keeps the base model frozen.]{\includegraphics[width=3.4in]{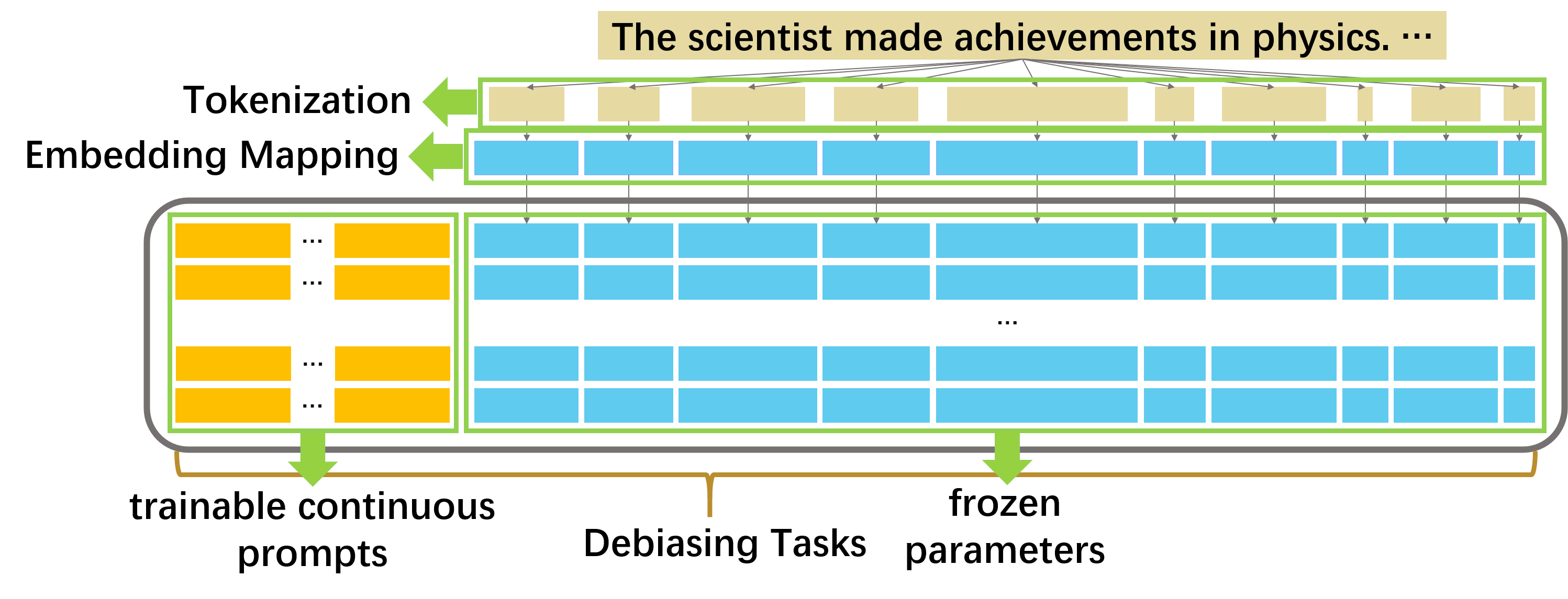} \label{1-a}}
    \hfill
    \subfloat[When performing downstream tasks, ADEPT conditions either the base model or both the prompt and the base model.]{\includegraphics[width=3.4in]{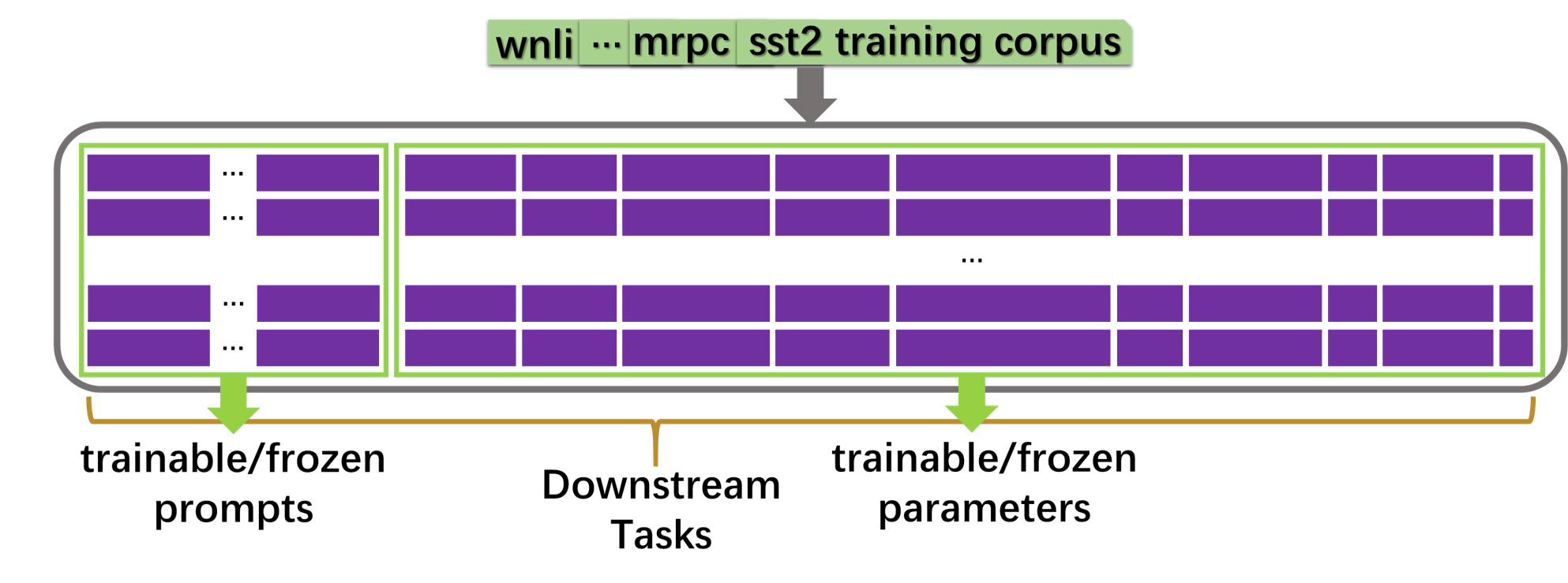} \label{1-b}}
    \caption{An illustration of how debiasing works using ADEPT and for downstream tasks.}
    \label{fig:1}
\end{figure*}

Natural Language Processing (NLP) tools are widely used today to perform reasoning and prediction by efficiently condensing the semantic meanings of a token, a sentence or a document. As more powerful NLP models have been developed, many real-world tasks have been automated by the application of these NLP systems. However, a great number of fields and tasks have a high demand for fairness and equality: legal information extraction \cite{rabelo2022overview}, resume filtering \cite{9799257}, and general language assistants \cite{askell2021general} to name a few. Unfortunately, in the pursuit of the most competitive results, folks often blindly apply PLMs, leading to strong performance with the unseen cost of introducing bias into the process. 
% As the concept and technology of word embeddings have been well developed, many efforts that regard language at its heart, \heng{' regard language at its heart' is akward} such as legal information extraction and entailment \cite{rabelo2022overview}, job resume filtering \cite{9799257}, and creating a general language assistant \cite{askell2021general}, are adopting the latest NLP tools to tackle their problems. In addition to the efforts tightly connected with law, recruitment, or general language assistants, there still exist a great number of fields and related tasks that have a high demand of fairness and justice. In other words, in most cases, 
An ideal NLP tool's decision or choice should not impose harms on a person based on their background \cite{blodgett2020language}, but many studies \cite{caliskan2017semantics,mayfield2019equity} have found that biases exist and occur throughout the NLP lifecycle. Thus, it is increasingly important that PLMs can be debiased to enable applications that may be inadvertently influenced by the PLM's implicit stereotypes.

% Some studies have disclosed the existence of biases in commonly used NLP tools, and as a result, many debiasing methods have emerged. The probe of biases and approaches to declining biases in static word embeddings have been studied for many years \cite{mikolov2013linguistic,NIPS2016_a486cd07,caliskan2017semantics,mikolov2013efficient,manzini2019black}. Compared with the static word embedding, the contextualized ones generated by PLMs takes the word's context into consideration, so a word may have a completely different mathematical representation based on the context in which it occurs. Due to this reason, it is generally believed that debiasing for contextualized word embeddings is more challenging than for static ones. Moreover, the debiasing of the former requires to make changes to a PLM's parameters, and after that some information learnt in the pretraining process may be eliminated, which often leads to the debiased model's performance degeneration in downstream tasks.

Debiasing, if treated as a special case of downstream tasks, can be tackled through finetuning. Typically, a finetuning debiasing method puts forward specific loss terms to guide a PLM to remove biases in itself \cite{kaneko2021debiasing}. Prompt tuning \cite{li2021prefix,liu2021gpt,lester2021power} is one of the more promising methods for transfer learning with large PLMs these days, and its general success \cite{raffel2020exploring} suggests applications toward debiasing as well. Prompt tuning, whose role is similar to that of finetuning, refers to freezing all the parameters of the original PLM and only training an additional section of parameters (called a ``prompt") for the downstream tasks. Here, a prompt is a set of tokens, often added as a prefix to the input for the task, that act as task-specific complementary information. 

All PLM debiasing methods must overcome a major hurdle of ``imbalance." Methods that are imbalanced do not adequately balance eliminating biases in a PLM while maintaining its representation ability. Some existing methods are prone to be destructive, whether destructive refers to decreasing a word/sentence embedding's projection on a linear bias subspace \cite{liang2020towards}, or refers to completely removing the semantic meanings of attribute words (e.g., man, male; and woman, female) from all neutral words (e.g., engineer, scientist; and teacher, librarian) \cite{kaneko2021debiasing}. 
% \heng{need example for attribute words and neutral words} 
If a debiasing framework focuses only on the PLM's debiasing task and pays no attention to preserving the model's useful properties, it may destroy the PLM's computational structure and counteract the benefits of pretraining altogether. Although an extreme example, a randomly initialized model is expected to be completely unbiased.

In this paper, we propose \textbf{ADEPT} (Figure \ref{fig:1}), a debiasing algorithm which implements prompt tuning to debias PLMs and makes the following contributions:

\begin{itemize}
    \item We are the first to exploit prompt tuning in the debiasing space.
    \item We introduce a novel debiasing criterion, which often enables the debiased model to perform better than the original one in downstream tasks.
    \item We show that \textbf{ADEPT} is more effective at mitigating biases on a word embedding manifold than other methods which operate on a linear bias subspace.
    % \item We delve into the corpus for an accurate prototype of a certain attribute on the manifold and set standards of evaluating the corpus regarding its reliability, quality, and quantity.
    \item We show methods for improving prototypes for contextualized word embeddings that are generated via aggregation.  
\end{itemize}

Our prompt tuning approach has the inherent advantage of saving computing and storage resources. In our experiments, we achieve great results by training prompts with less than 1\% the parameters of the PLM as opposed to fine-tuning approaches which train the whole model. Furthermore, because prompt tuning only trains prompt and the PLM's original parameters are not touched during the training process, the base model will maintain its robustness. 
\section{Related Work}

% \heng{this section gives a good summary of related work, but lacks of comments on their pros and cons, and also lacks of comparison with your work}

\subsection{Debiasing Methods}
\subsubsection{Word Embeddings}
Static word embeddings, the foundational building blocks of neural language models, have been a prime target for criticism. In light of artifacts from their training process leading to the encoding of stereotypes, many efforts have been made to mitigate the correlations stored within static embeddings \citep{mikolov2013linguistic,NIPS2016_a486cd07,caliskan2017semantics,mikolov2013efficient,manzini2019black}. However, most modern PLMs employ contextualized word embeddings, spreading the potentially biased representations of words across various contexts. 

\subsubsection{Discrete Prompts}

% \heng{You did a good job at describing what you did, but each of these steps needs some story telling about why you designed methods in this way, what was the intuition, etc.}
\citet{solaiman2021process} propose PALMS with Values-Targeted Datasets, which finetunes large-scaled PLMs on a predetermined set of social values in order to reduce PLMs' biases and toxicity. \citet{askell2021general} use a hand-designed prompt with more than 4600 solid words as a stronger baseline for helpfulness, harmlessness, and honesty principle for a general language assistant. \citet{schick2021self} encourage a model to generate biased text and discard its undesired behaviors with this internal knowledge. 

In general, discrete prompts debias PLMs in the form of debiasing descriptions. As crafting discrete prompts manually requires domain knowledge and professional expertise, and we cannot ensure hand-crafted prompts' effectiveness beforehand, we hope to improve debiasing prompts' performance by transforming it to continuous ones which can be optimized with standard techniques like gradient descent.%Similarly, \citet{guo-etal-2022-auto} search biased prompts in a PLM and use these prompts to mitigate biases. 
% \subsubsection{SENT-DEBIAS}
% \citet{liang2020towards} put forward SENT-DEBIAS for debiasing sentence representations, which is the sentence-level extension of \citet{NIPS2016_a486cd07}'s research. The work makes an assumption that there exists a linear bias subspace in the sentence representation, constructs the subspace with contextualized bias attributes, and removes the biases by reducing the sentences' projection on that subspace.

\subsubsection{Finetuning Setting}
\citet{kaneko2021debiasing} propose a finetuning method of debiasing PLMs. It sets special loss for the debiasing tasks which takes both a PLM's debiasing results and its expressiveness into account. The experiment shows that token-level debiasing across all layers of the PLM produces the best performance. It further conducts experiments on MNLI tasks and finds that the debiased model preserves semantic information. As this work also makes efforts to maintain a PLM's expressiveness while debiasing, we take their debiased model as our baseline.
\subsection{Prompt Tuning}
Prompt usually has two connotations. One is the text with natural semantics, which is fed into the language model together with the original input as additional information. Another is a set of prefixed, continuous trainable numbers post-set into a PLM, which usually do not have semantic meanings. Because this set of continuous numbers have the same functions as the discrete prompt, such as providing the PLM with extra hints for solving a problem, it is also called prompt (or prefix).

\citet{li2021prefix}, \citet{liu2021gpt}, and \citet{lester2021power} propose prompt tuning (or prefix-tuning, p-tuning) as a lightweight alternative to finetuning for performing downstream tasks. This approach conditions a large-scaled PLM by freezing its original parameters and optimizing a small continuous task-specific embeddings. Besides saving computing and storage resources, prompt tuning performs even better when the PLM scales up and keeps the PLM's robustness to domain transfer. Our work benefits from these advantages as we debias large PLMs and evaluate their expressiveness on downstream tasks.

\subsection{Manifold Learning}
Manifold learning refers to a series of machine learning methods based on manifold assumption \cite{melas2020mathematical}. In order to grasp knowledge from data, we need to hypothesize that data has its inborn structure. Manifold assumption indicates that the observed data lie on a low-dimensional manifold embedded in a higher-dimensional space, for example, a Swiss Roll alike data structure in a 3-dimensional data space. t-SNE \cite{van2008visualizing}, a decomposition method based on manifold assumption, provides excellent visualizations for high-dimensional data that lie on several different, but related, low-dimensional manifolds. For word embeddings with high dimensions, we believe we can better describe its distribution with a manifold than with a linear subspace.

\section{Methodology}
\begin{algorithm}[t!]
\caption{\textbf{ADEPT}: a debiasing algorithm for contextualized word embeddings.}
\label{alg:ADEPT}
\textbf{Input}: a Pre-trained Language Model (PLM)\\
\textbf{Output}: $\Phi_{prompt}$ for debiasing the PLM\\
\textbf{ADEPT}:
\begin{algorithmic}[1] %[1] enables line numbers
\STATE Prepare a PLM $M_{\Theta}$ with parameters $\Theta$.
\STATE Suppose a bias has $d$ attributes. Define a neutral word tuple $W^{neutral}$ and attribute word tuples $W^{a(i)}=(w_1^{a(i)}, ..., w_g^{a(i)})$, each with $g$ one-to-one words.
\STATE Collect sentences $S^{neutral}$ and $\{S^{a(i)}\}_{i=1}^d$.
\STATE Initialize parameters $\Phi_{prompt}$.
\FOR {epoch in 1, ..., $epoch_{max}$}
\STATE Calculate prototypes of the neutral words:
\\$E^{neutral}=M_{\Theta}^{'}(S^{neutral})$,
\\where $M_{\Theta}^{'}=M_{\Theta \cup \Phi_{prompt}}$.
\STATE Calculate prototypes of attributes: 
\\$E^{a(i)}=M_{\Theta}^{'}(S^{a(i)}),e^{a(i)}=aver(E^{a(i)})$.
\STATE Calculate distances between attribute words and neutral words: $P^{a(i)}=Distance(E^{neutral}|e^{a(i)})$.
\STATE Calculate loss of bias:
\\$L_{bias}=\sum_{i,j\in \{1, ..., d\}, i<j}\{JS(P^{a(i)}||P^{a(j)})\}$.
\STATE Calculate loss of representation:
\\$L_{representation}=KL(M_{\Theta}(S)||M_{\Theta}^{'}(S))$,
\\where $S=S^{neutral}\cup \{S^{a(i)}\}_{i=1}^d$.
\STATE Calculate the total loss:
\\$L=L_{bias}+\lambda L_{representation}$.
\STATE Compute gradient.
\STATE Update $\Phi_{prompt}$.
\ENDFOR
\STATE \textbf{return} best $\Phi_{prompt}$
\end{algorithmic}
\end{algorithm}
Our goal is: given a PLM $M_{\Theta}$ with parameters $\Theta$, find the parameters $\Phi_{prompt}$ determining a set of continuous prompts, so that the prompt-tuned model $M_{\Theta \cup \Phi_{prompt}}$ (we will use $M_{\Theta}^{'}$ for short) has the debiasing effects while maintaining the expressiveness of $M_{\Theta}$.

We optimize $\Phi_{prompt}$ by using the objective function: 
\begin{equation}\label{eqa:total loss}
    L=L_{bias}+\lambda L_{representation}
\end{equation}
where $L_{bias}$ seeks to minimize biases in $M_{\Theta}^{'}$ whereas $L_{representation}$ caters to the debiased model's expressiveness, and $\lambda$ is a coefficient to balance the two dependent terms. Our algorithm is summarized in Algorithm \ref{alg:ADEPT}.

\subsection{Define Word Tuples and Collect Sentences}
We define a neutral word tuple $W^{neutral}$ and several attribute word tuples $W^{a(i)},i=1, ..., d$, where the category of bias we are debiasing for contains $d$ different attributes. For example, gender bias may have the attributes ``female" and ``male," and here $d=2$ \footnote{We hold the opinion that gender identity need not be restricted to the binary choice of male or female. However, for the purposes of experimentation and following prior studies, we adopt this binary setting.}. Words in $W^{neutral}$ are nouns or adjectives that should show no preference for any of the $d$ attributes. For example, ``science" and ``ambitious," which should not be bound to any attributes, might be in the tuple $W^{neutral}$. $W^{a(i)}$ denotes a tuple of words where each word is associated attribute $a(i)$ and not $a(j)$ for any $j\neq i$. For example, $W^{male}$ might contain the words ``uncle" and ``masculine" but not contain the word ``science" (since that is a neutral word) or the word ``parent" (as this is not specific to the ``male" attribute). Next, we enforce that each attribute word tuple is indexed by the same (implicit) indexing set of \textit{concepts} and that the word at each index is of the same form. For example, if the (implicit) indexing set is (``parent's sibling", ``parent", ``sibling") then the male attribute tuple would be (``uncle", ``father", ``brother") and the female attribute tuple would be (``aunt", ``mother", ``sister"). For brevity, we use the word ``pairwise" to describe this correspondence, although the method can be extended to biases with $d>2$ as well. 
% We have to mention that words in $W^{a(i)}=(w_1^{a(i)}, ..., w_g^{a(i)})$ have sequential one-to-one correspondence across all $d$ attributes. For example, ``aunt" (or ``feminine") lies in $W^{female}$ with the same index as ``uncle" (or ``masculine") in $W^{male}$. For brevity, we will use ``pairwise" for describing the property ``with one-to-one correspondence," though the method is extensible to biases with any attribute number. 

We then collect sentences based on the word tuples. $S^{neutral}$ (or $S^{a(i)}$) denotes sentences that contain at least one word in $W^{neutral}$ (or $W^{a(i)}$, respectively). Instead of creating template-based sentences using the attribute words from $\{W^{neutral}\}\cup \{W^{a(i)}\}_{i=1}^d$, we scrape natural sentences from a corpus (possibly distinct from and/or smaller than the PLM's pretraining corpus) for a diverse word distribution that aligns better with the real-world.

\subsection{Calculate Prototypes of Neutral Words/Attributes}
To get an insight of a model's view on different groups, we seek prototypes of neutral words and attributes. To obtain these prototypes, we extract embeddings for each word. For a word from $W^x$ ($x$ is $neutral$ or $a(i)$ for some $i$), we fetch the associated sentence from $S^x$ and feed it into $M_{\Theta}^{'}$. Then, we extract the hidden state for the word from each layer of the forward pass. For PLMs adopting WordPiece embeddings such as BERT \citep{devlin2018bert}, if a word has several sub-tokens, we average the sub-tokens' hidden states as the word's hidden state. 

For each word tuple's sentences $S^{x}$, we extract the set of embeddings $E^{x}$. For attribute words, we follow the procedures from \citet{bommasani-etal-2020-interpreting} and average the embeddings $E^{a(i)}$ to get a single embedding $e^{a(i)}$ that closer resembles a static embedding as opposed to contextualized embeddings. Under the law of large numbers, we expect this simple linear computation to reduce the context's linear influences on each attribute word. This process can be summarized as: 
\begin{equation}\label{equ:prototype list}
\begin{aligned}
    E^{neutral}&=M_{\Theta}^{'}(S^{neutral}) \\
    E^{a(i)}&=M_{\Theta}^{'}(S^{a(i)})
\end{aligned}
\end{equation}
\begin{equation}\label{equ:prototype}
    e^{a(i)}=aver(E^{a(i)})
\end{equation}
Thus we take $E^{neutral}=[e^{neutral}_1, e^{neutral}_2, ...]$ as the prototypes of neutral words and $e^{a(i)}$ as the prototype of an attribute.

\subsection{Define Tuning Loss}
We treat word embeddings as being distributed on a manifold and design the loss adhering to the criterion that pairwise attribute words should look alike compared to neutral words on the manifold.

We first design $L_{bias}$ with the intention of pushing pairwise attribute words closer together on the manifold, which corresponds to decreasing biases in a PLM. $p_{n_j|a(i)}$ quantifies the degree to which attribute $a(i)$'s information can be restored from the neutral word $n_j$ in $M_{\Theta}^{'}$:\\
\begin{equation}\label{equ:p}
    p_{n_j|a(i)}=\frac{exp(-\frac{||e^{a(i)}-e^{neutral}_{j}||^2}{2\rho^2})}{\sum_{n_k\in W^{neutral}}\{exp(-\frac{||e^{a(i)}-e^{neutral}_{k}||^2}{2\rho^2})\}}
\end{equation}
where $\rho$ is a hyperparameter. We can interpret Equation \ref{equ:p} in this way: (1) Let us set a Gaussian distribution with a covariance matrix to be $\rho$ times the identity matrix at the prototype of attribute $a(i)$, which is $e^{a(i)}$. Then the prototype of the neutral word $n_j$, which is $e^{neutral}_j$, shows up in the distribution with the probability proportional to $exp(-\frac{||e^{a(i)}-e^{neutral}_{j}||^2}{2\rho^2})$, the numerator. (2) The denominator sums up the probability mentioned above from all $n_k\in W^{neutral}$, and plays a role as the normalization factor. (3) Equation \ref{equ:p} is a formulation that quantifies how much information of $e^{a(i)}$ we can restore from $e^{neutral}_j$. Similar equations have been used in other contexts \citep{10.2307/2237880, hinton2002stochastic}.

$P^{a(i)}$ denotes our distances from attribute $a(i)$ to all neutral words. $P^{a(i)}=[p_{n_1|a(i)}, p_{n_2|a(i)}, ...]$ means it is a list of values calculated from $e^{a(i)}$ and $E^{neutral}$. Therefore, we summarize it as below:
\begin{equation}\label{equ:P}
    P^{a(i)}=Distance(E^{neutral}|e^{a(i)})
\end{equation}

We define $L_{bias}$ as:
\begin{equation}\label{equ:bias loss}
    L_{bias}=\sum_{i,j\in \{1, ..., d\}, i<j}\{JS(P^{a(i)}||P^{a(j)})\}
\end{equation}
where $JS(P^{a(i)}||P^{a(j)})$ is the Jensen-Shannon divergence between distribution $P^{a(i)}$ and distribution $P^{a(j)}$. This loss term is intended to make up the difference between distinct attributes' relative distances to the same group of neutral words (in the form of distribution $P^{a(i)}$ and $P^{a(j)}$) so as to push pairwise attribute words closer.

We then design $L_{representation}$ with the intention of maintaining words' relative distances, which corresponds to maintaining the PLM's representation ability. $p_{w_j|w_i}$ quantifies the degree to which the word $w_i$'s information can be restored from the word $w_j$ in $M_{\Theta}^{'}$. $P$ denotes the matrix of $p_{w_j|w_i}$ where $P_{ij}=p_{w_j|w_i}$. For $q_{w_j|w_i}$ and $Q$, they denote likewise except that the model is the original one $M_{\Theta}$. $p_{w_j|w_i}$ and $q_{w_j|w_i}$ have the same definition as in Equation \ref{equ:p}.

We define $L_{representation}$ as:
\begin{equation}\label{equ:representation loss}
\begin{aligned}
    L_{representation}&=KL(Q||P) \\
    &=\sum_{i=1}^{||V||}\sum_{j=1}^{||V||}Q_{ij}log_2(\frac{Q_{ij}}{P_{ij}})
\end{aligned}
\end{equation}
where $||V||$ denotes vocabulary size.

In Algorithm \ref{alg:ADEPT}, we write $L_{representation}$ as:
\begin{equation}\label{equ:representation loss in use}
    L_{representation}=KL(M_{\Theta}(S)||M_{\Theta}^{'}(S))
\end{equation}
where S denotes the union of $S^{neutral}$ and $\{S^{a(i)}\}_{i=1}^d$. Here the $L_{representation}$ aims to keep the PLM’s parameters unchanged. Rather than using $L_2 $ norm to gauge how much the outputs of the debiased model has changed as \citet{kaneko2021debiasing} do, we measure the differential between the original model's hidden states and the debiased model's hidden states with KL divergence. $L_{representation}$ in Equation \ref{equ:representation loss in use} is more time-efficient for training and evaluation tasks than the one in Equation \ref{equ:representation loss}, so we adopt it in \textbf{ADEPT}. 

\subsection{Improve Prototypes of Attributes}
After we confirm that Algorithm \ref{alg:ADEPT} works, we make efforts to improve prototypes of attributes $e^{a(i)}$ by adjusting properties of $S^{a(i)}$. We can tell from Equation \ref{equ:prototype} that $e^{a(i)}$ is a calculated prototype with intuitive correctness. Therefore, we implement experiments on deciding on the desirable properties of $S^{a(i)}$ regarding its reliability, quality and quantity, altering single variable at a time, to check whether $e^{a(i)}$'s expressiveness can be improved with the modified $S^{a(i)}$. The test granularity extends from a single word to the whole attribute.

$S^{a(i)}_m$ denotes a sub-list of $S^{a(i)}$ composed with sentences that contain $w^{a(i)}_m$, where $w^{a(i)}_m$ means the $m^{th}$ item of tuple $W^{a(i)}$. $len(S^{a(i)}_m)$ and $len(S^{a(i)})$ denote the length of the lists.
\subsubsection{Reliability}
Here, the experiment is devised to answer: if $len(S^{a(i)}_m)$ is less than a threshold, shall we take the word $w^{a(i)}_m$ as a contributing word for constructing $e^{a(i)}$? To satisfy the law of large number, we set the threshold to be 30.
\subsubsection{Quality}
Here, the experiment is devised to answer: if $len(S^{a(1)}_m)\neq len(S^{a(2)}_m)\neq ...$, which is often the case, will this disproportion of pairwise words affect $e^{a(i)}$'s expressiveness? We set $len(S^{a(1)}_m)=len(S^{a(2)}_m)=..., m\in [1, g]$ and compare the results.
\subsubsection{Quantity}
Here, the experiment is devised to answer: whether for $len(S^{a(i)})$, the larger, the better? We conduct the tests with the $len(S^{a(i)})$ being of a different order of magnitude.
\section{Experiments}
\subsection{Datasets, Benchmarks and Baselines}
For the word tuples, we use neutral word lists employed in previous debiasing methods \citep{kaneko2021debiasing, caliskan2017semantics}. For the binary gender setting, we use the pairwise attribute words from \citet{zhao2018learning} and for the ternary religion setting, we use the attribute triplets from \citet{liang2020towards}. For the sentences associated with the word tuples, we draw sentences from News-Commentary v15 \cite{TIEDEMANN12.463} for the gender setting and sentences from BookCorpus \cite{Zhu_2015_ICCV} and News-Commentary v15 \cite{TIEDEMANN12.463} for the relgions setting. Since the original BookCorpus is no longer available, we use \cite{bookcorpus_hf} which is an open source replica. We use this corpus since BookCorpus is part of the corpus BERT is originally trained on. In total, for the gender setting, we draw 20,710 neutral sentences and 44,683 sentences each for the male and female attributes. For the religion setting, we draw 73,438 neutral sentences and 5,972 sentences corresponding to each attribute (Judaism, Christianity, and Islam). 

We evaluate gender stereotype scores on SEAT 6, 7, 8 \cite{may2019measuring} and CrowS-Pairs \cite{nangia2020crows}, widely-used benchmarks/metrics designed to evaluate a model's biases toward/against different social groups. We evaluate the debiased models' representation ability on selected GLUE \cite{wang2018glue} tasks, each of which is with little training data: Stanford Sentiment Treebank (SST-2, \citep{socher-etal-2013-recursive}), Microsoft Research Paraphrase Corpus (MRPC, \citep{dolan-brockett-2005-automatically}) Recognizing Textual Entailment (RTE, \citep{dagan2005pascal,haim2006second,giampiccolo2007third,bentivogli2009fifth}) and Winograd Schema Challenge (WNLI, \citep{levesque2012winograd}). We further evaluate the comprehensive performance of the debiased model pertaining to its biases and expressiveness on a filtered portion of the StereoSet-Intrasentence data \citep{nadeem2020stereoset}, employing 149 test examples for the gender domain and 5,770 test examples overall.

We compare our algorithm with Debiasing Pre-trained Contextualised Embeddings (\textbf{DPCE}; \citep{kaneko2021debiasing}), a similar method that focuses on making the neutral words' embeddings devoid of information in relation to a protected attribute by finetuning the model with the loss term being the sum of the inner product between the attribute words' hidden states and the neutral words' hidden states.

\subsection{Hyperparameters}
We conduct experiments on the \textsc{bert-large-uncased} pre-trained model from HuggingFace \citep{wolf2019huggingface}. By using \textbf{ADEPT}, we need only train 1.97M parameters when prompt-tuning with 40 prompt tokens, orders of magnitude smaller than the 335M parameters required for finetuning. 

We set $\lambda$ in Equation \ref{eqa:total loss} to be $\frac{7}{3}$ and $\rho$ in Equation \ref{equ:p} to be 15. We use Adam \cite{kingma2014adam} to optimize the objective function. During the debiasing process, our learning rate is 5e-5 and our batchsize is 32. Results for \textbf{DPCE} are using the hyperparameters originally reported in \citet{kaneko2021debiasing}. All the experiments are conducted on two GeForce RTX 3090 GPUs and in a Linux operating system.

\subsection{Bias Benchmarks}
We use three main benchmarks for evaluating performance vis-a-vis bias. 

\subsubsection{SEAT}
The Sentence Encoder Association Test (SEAT) \citep{may2019measuring} extends the Word-Embedding Association
Test (WEAT) \citep{caliskan2017semantics} to the sentence-level by filling hand-crafted templates with the words in WEAT. In this way, SEAT aims to measure biases in sentence-encoders like ELMo \citep{peters-etal-2018-deep} and BERT \citep{devlin2018bert} as opposed to only the biases in word embeddings. The SEAT benchmark provides two scores, namely effect size and P-value, where an effect size with smaller absolute value is regarded as a better score for a debiased model.

\subsubsection{CrowS-Pairs}
CrowS-Pairs \citep{nangia2020crows} features pairwise test sentences, differing only in a stereotyped word and an anti-stereotyped word in the same position. This benchmark evaluates whether a PLM will assign a higher probability to a stereotyped sentence than to an anti-stereotyped one where the probability is assigned while attempting to account for differing priors. A ideal model will get the score of 50. 

\subsubsection{StereoSet}
StereoSet \citep{nadeem2020stereoset} measures both a PLM's useful semantic information as well as its biases by using cloze tests. Provided a brief context, a PLM must choose its preference from a stereotype, an anti-stereotype, and an unrelated choice. A higher, up to 100, Language Modeling Score (LMS) indicates better expressiveness, and a Stereotype Score (SS) closer to 50 indicates less biases. The Idealized CAT Score (ICAT) is a combined score of LMS and SS with the best score being 100. % For both CrowS-Pairs and StereoSet, as they require the model to predict masked contents during the evaluation, a debiased model is at a disadvantage compared to the original one when making predictions with a pretrained classifier.

\subsection{Results and Analysis}

% \heng{Table 2 is very hard to understand for people who are not familiar with the task. I suggest you ton introduce some background, and you can move some details on the parameters to the appendix.}

\begin{table*}[t!]
\renewcommand{\arraystretch}{1.2}
\centering
\begin{tabular}{l|c|c|c|cl}
\hline
                                  & \textbf{original} & \textbf{DPCE}   & \textbf{ADEPT-finetuning} & \multicolumn{2}{c}{\textbf{ADEPT}}                 \\ \hline
C6: M/F Names, Career/Family      & 0.369             & 0.936           & 0.328                     & \multicolumn{2}{c}{\textbf{0.120}}                 \\
C7: M/F Terms, Math/Arts          & 0.418             & -0.812          & \textbf{-0.270}           & \multicolumn{2}{c}{-0.571}                         \\
C8: M/F Terms, Science/Arts       & -0.259            & -0.938          & -0.140                    & \multicolumn{2}{c}{\textbf{0.132}}                 \\ \hline
CrowS-Pairs: score(S)             & 55.73             & 47.71           & 52.29                     & \multicolumn{2}{c}{\textbf{48.85}}                 \\ \hline
GLUE: SST-2                       & 92.8              & 92.8            & \textbf{93.6}             & \multicolumn{1}{c|}{93.3}          & 92.7          \\
GLUE: MRPC                        & 83.1              & 70.3            & 83.6                      & \multicolumn{1}{c|}{84.6}          & \textbf{85.0} \\
GLUE: RTE                         & 69.3              & 61.0            & 69.0                      & \multicolumn{1}{c|}{\textbf{69.7}} & \textbf{69.7} \\
GLUE: WNLI                        & 53.5              & 45.1            & 46.5                      & \multicolumn{1}{c|}{47.9}          & \textbf{56.3} \\ \hline
StereoSet(filtered)-gender: LMS   & \textbf{86.338}   & 84.420           & 86.005                    & \multicolumn{2}{c}{84.652}                         \\
StereoSet(filtered)-gender: SS    & 59.657            & 59.657          & 57.113                    & \multicolumn{2}{c}{\textbf{56.019}}                \\
StereoSet(filtered)-gender: ICAT  & 69.663            & 68.115          & 73.770                    & \multicolumn{2}{c}{\textbf{74.462}}                \\
StereoSet(filtered)-overall: LMS  & 84.162            & 58.044          & \textbf{84.424}           & \multicolumn{2}{c}{83.875}                         \\
StereoSet(filtered)-overall: SS   & 58.243            & \textbf{51.498} & 57.701                    & \multicolumn{2}{c}{55.435}                         \\
StereoSet(filtered)-overall: ICAT & 70.288            & 56.305          & 71.420                    & \multicolumn{2}{c}{\textbf{74.759}}                \\ \hline
\end{tabular}
\caption{Evaluation results on debiasing performance. We test the debiased models on SEAT (from row 1 to 3), CrowS-Pairs (row 4), GLUE (from row 5 to 8) and filtered StereoSet-Intrasentence (from row 9 to 14), with best result in bold. original, the original \textsc{bert-large-uncased} model; DPCE \cite{kaneko2021debiasing}, the baseline model; ADEPT-finetuning, model finetuned with our new debiasing criterion; ADEPT, model tuned with ADEPT. As for downstream tasks in GLUE, we test ADEPT on both fine-tuning the original model only (left) and fine-tuning the model as well as the debiasing prompt (right).}
\label{tbl:experiments}
\end{table*}
We evaluate four models on all benchmarks, namely the \textbf{original} model (pre-trained with no explicit debiasing), the \textbf{DPCE} model, the \textbf{ADEPT-finetuning} model finetuned following our debiasing criterion, and the \textbf{ADEPT} model (ours). For the CrowS-Pairs and StereoSet experiments, we inherit the classifier from the \textsc{bert-large-uncased} model to predict the masked token, so we perform CrowS-Pairs and StereoSet evaluations on the model with the slightest change. As a result, we choose \textbf{ADEPT} model with 500 training steps in these two benchmarks' evaluation. For SEAT and GLUE, we use the \textbf{ADEPT} model after 10 epochs of training.

\subsubsection{Reducing Biases}
In Table \ref{tbl:experiments}, experiments show that \textbf{ADEPT} achieves competitive debiasing results, outperforming \textbf{DPCE} and mostly obtaining the best scores of the four models on SEAT and CrowS-Pairs. \textbf{ADEPT-finetuning}, which shifts \textbf{ADEPT} from the prompt-tuning setting to a finetuning setting, is also broadly successful at eliminates biases in PLMs. More trainable parameters notwithstanding, \textbf{ADEPT-finetuning} fails to beat \textbf{ADEPT}, implying that debiasing does not require a great change to the original model. 

\subsubsection{Preserving Representation Ability}
In Table \ref{tbl:experiments}, the GLUE tests show that \textbf{ADEPT} does not harm the model's representation ability and even improves it in most cases, with increased scores on SST-2, MRPC, and RTE. As shown in Figure \ref{fig:1}\subref{1-b}, after being debiased with \textbf{ADEPT}, the model can choose from training the base model or both the prompt and the base model when performing downstream tasks, and we test both on selected GLUE tasks. Notably, the trainable prompt parameters account for less than 1/100 of the base model parameters, so additionally training the prompts does not significantly add to the computational burden. We can see enhanced performance in both cases. \textbf{ADEPT-finetuning} also manages to outperform the \textbf{original} pre-trained model on SST-2 and MRPC, although the overall results are less noteworthy. 

\subsubsection{Visualization and Comprehensive Performance}
\begin{figure}[t!]
    \centering
    \subfloat[original]{\includegraphics[width=1.62in]{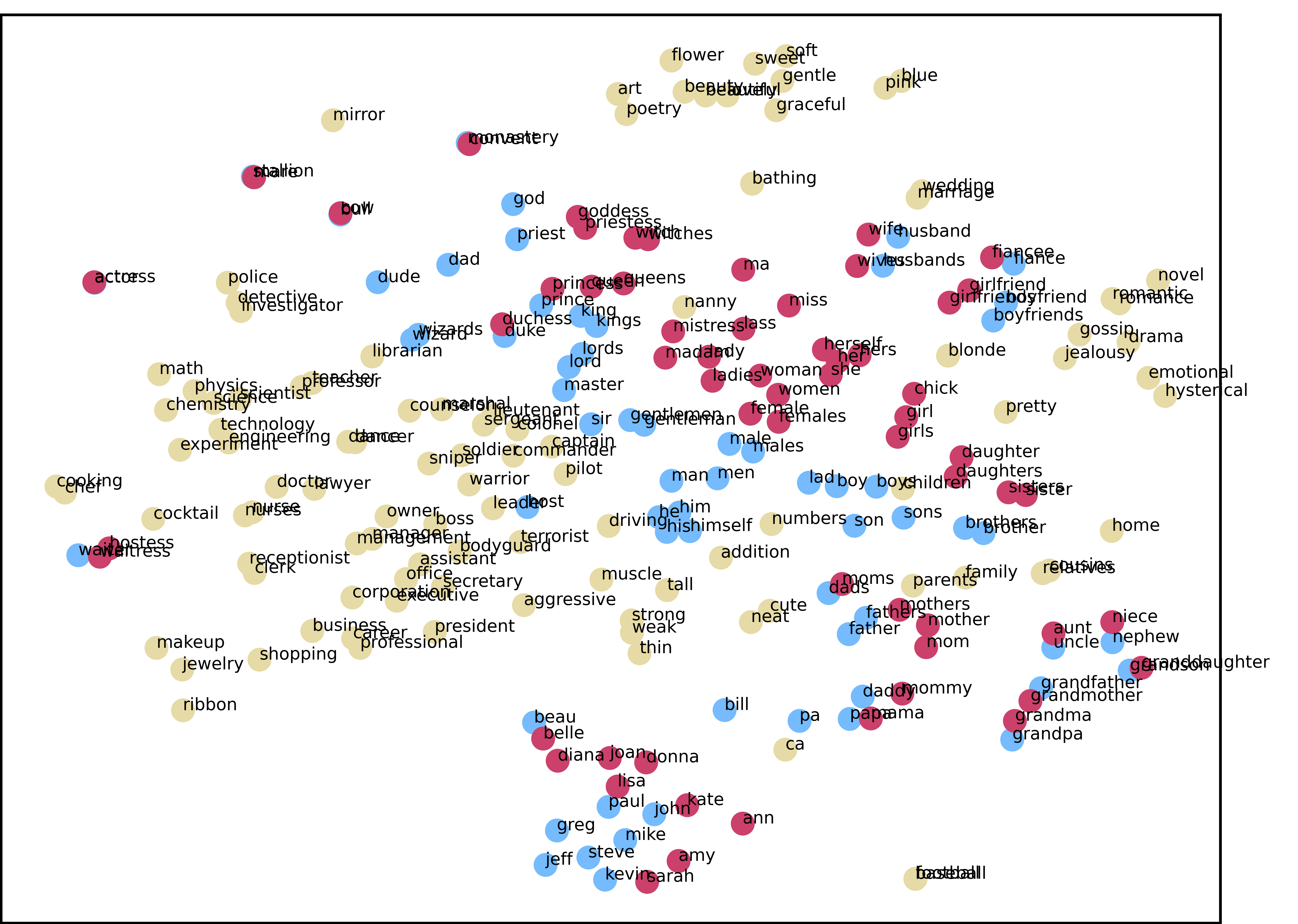} \label{2-a}}
    \hfill
    \subfloat[DPCE]{\includegraphics[width=1.62in]{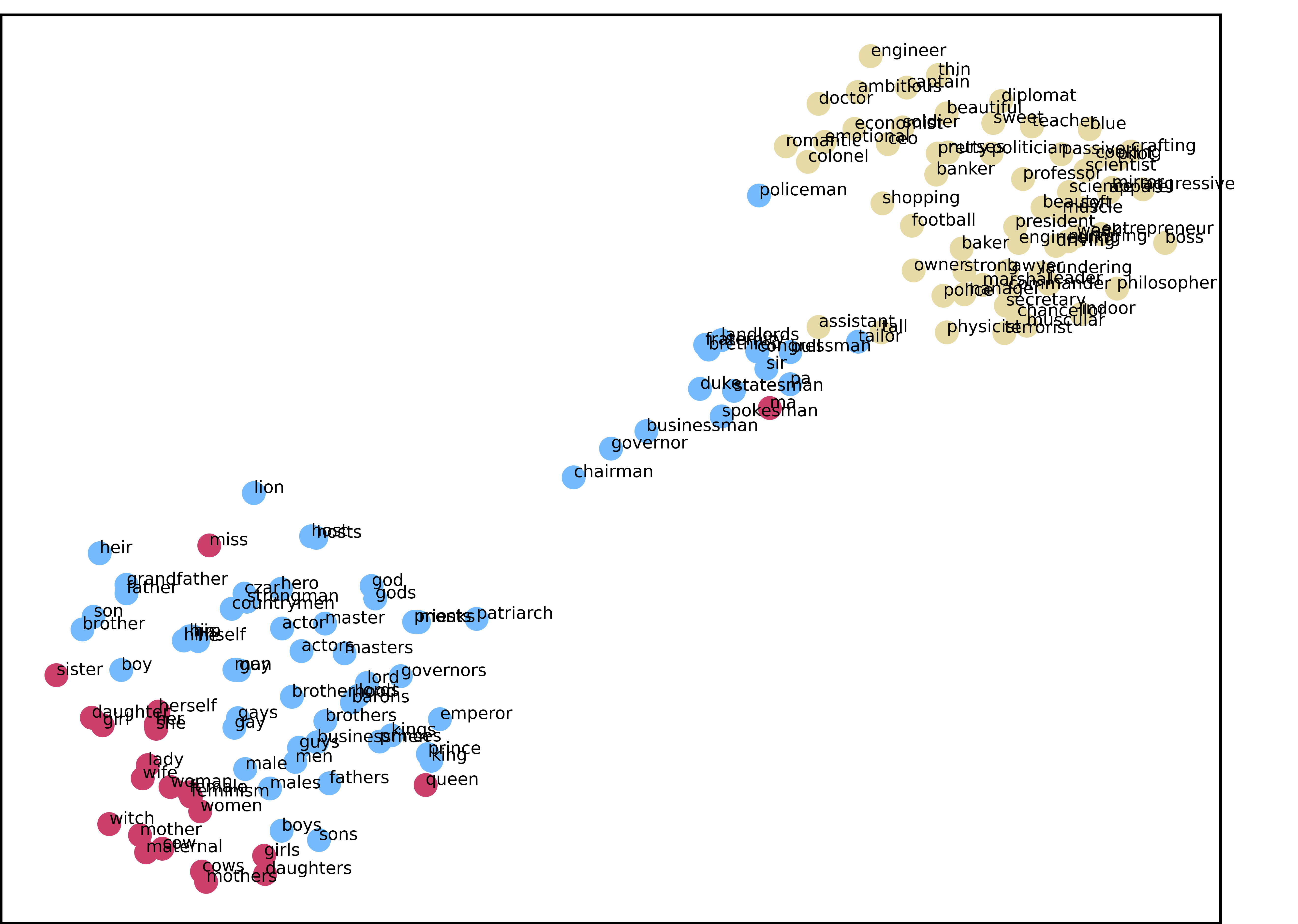} \label{2-b}}
    \hfill
    \subfloat[ADEPT-finetuning]{\includegraphics[width=1.62in]{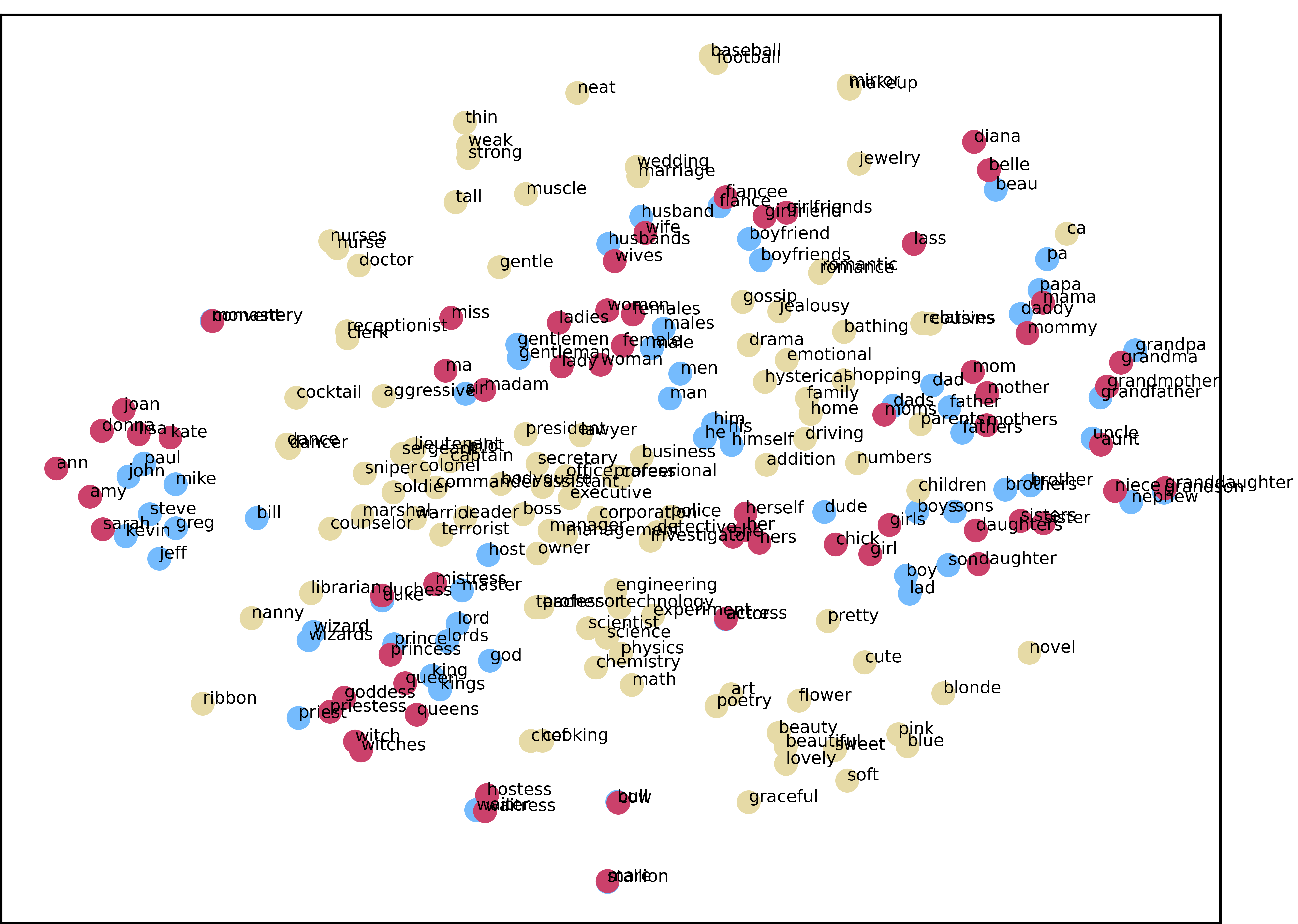} \label{2-c}}
    \hfill
    \subfloat[ADEPT]{\includegraphics[width=1.62in]{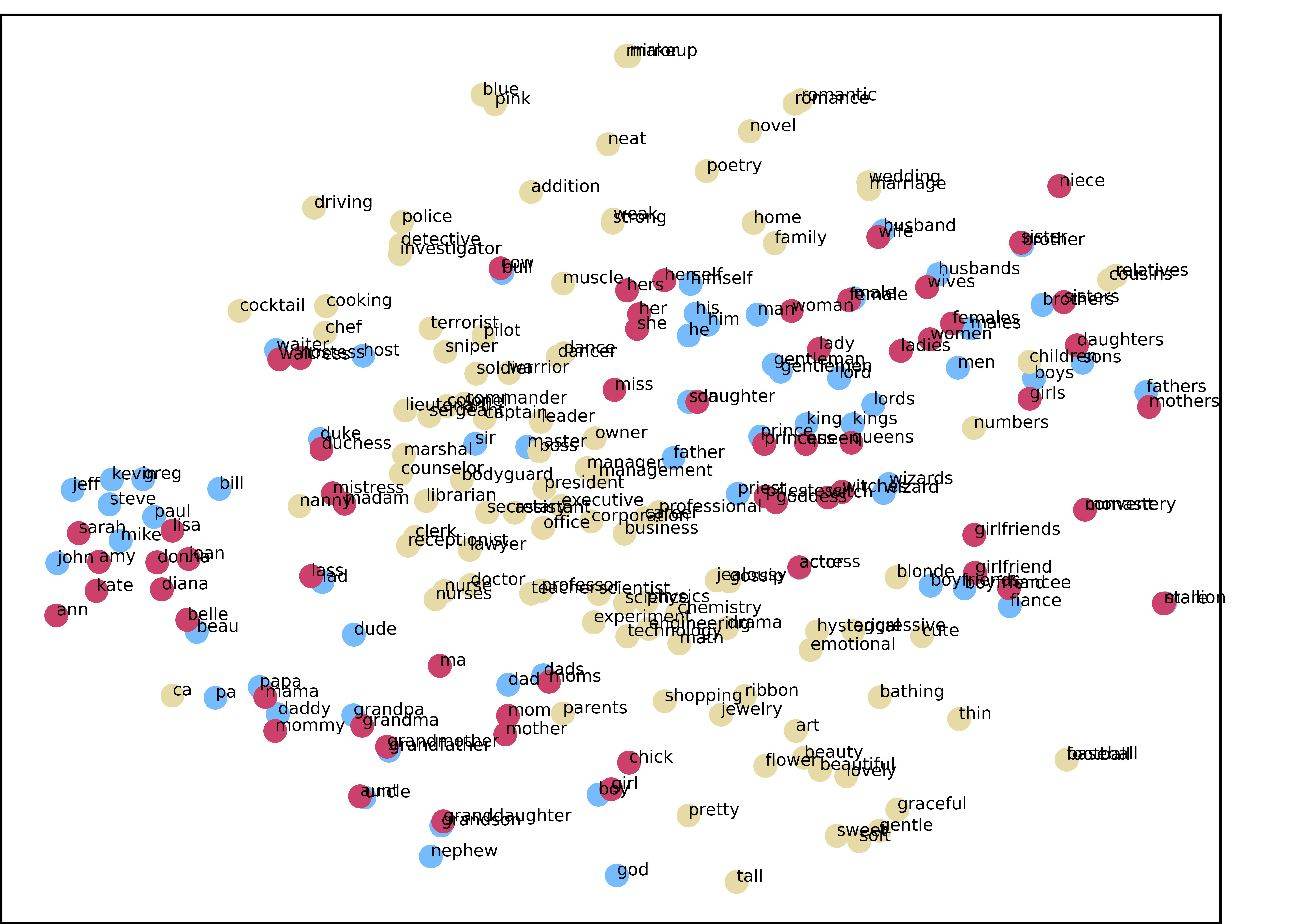} \label{2-d}}
    \caption{Visualized correlation of words in the gender domain. We use t-SNE to plot the figures and set perplexity as 30. We color neutral words beige, male words blue, and female words red. 
    % \textbf{original}, the original model; \textbf{DPCE} \cite{kaneko2021debiasing}, the baseline model; \textbf{ADEPT-finetuning}, model finetuned with our new debiasing criterion; \textbf{ADEPT}, model tuned with \textbf{ADEPT}. 
    }
    \label{fig:2}
\end{figure}

We explore the visible increase in the debiased model's expressiveness by visualizing words' correlation given by the model before and after debiasing, and provide a comprehensive score on the filtered StereoSet-Intrasentence. For a better prototype of a word, we average the last layer hidden state of the word from 30 different sentences. We plot \textbf{ADEPT}'s performance on binary gender debiasing in Figure \ref{fig:2}\subref{2-d} (we also plot the results for ternary religion debiasing in the Appendix). We filter the StereoSet-Intrasentence dataset to only keep test examples with the target words ``daddy," ``ma'am," ``groom," ``bride," ``stepfather," or ``stepmother,"  as these words show up less often in the tuning corpus. 

From Figure \ref{fig:2} we can conclude that \textbf{ADEPT} succeeds at maintaining words' relative distances, while simulatenously pulling pairwise attribute words closer. In comparison, as shown in Figure \ref{fig:2}\subref{2-b}, removing attribute semantic meanings as done in \textbf{DPCE} splits neutral words and gender words apart, which actually makes the difference between pairwise gender words negligible compared to their relative distances to the neutral words group. This may account for why some previous debiasing methods see a drastic drop in their model's expressiveness after debiasing whereas \textbf{ADEPT} does not. We further plot the evaluation loss in the training process in Appendix and find that the training process of \textbf{ADEPT} is smoother than that of \textbf{DPCE}.

The filtered StereoSet-Intrasentence result also implies that \textbf{ADEPT} is better at keeping useful semantic information when eliminating biases. \textbf{ADEPT} achieves the best ICAT score across the evaluated models in the filtered StereoSet-Intrasentence for gender and for overall, with best SS in gender domain and best LMS across domains. The baseline \textbf{DPCE} model appears to misunderstand words other than gender words as its LMS declines from 84 to 58 when the StereoSet-Intrasentence extends its examples from gender to other protected groups like race and religion. We note that the SS does not improve as much as the SEAT or CrowS-Pairs scores do. We hypothesize that this is because our training process is more similar to the metrics used for SEAT and CrowS-Pairs, which are calculated across the full sentence, rather than to StereoSet, which computes only for the word.

\subsection{Experiments for Improving Prototypes of Attributes}\label{sec:improving_prototypes}
\begin{table}[t]
\renewcommand{\arraystretch}{1.2}
\centering
\begin{tabular}{l|c|c|c|c}
\hline
                        & LMS    & SS     & ICAT   & score(S) \\ \hline
\textbf{raw}            & 86.674 & 62.341 & 65.282 & 52.29    \\
\textbf{reliability}    & 85.975 & 61.846 & 65.605 & 53.05    \\
\textbf{quality}        & 86.728 & 62.329 & 65.343 & 53.44    \\
\textbf{quantity-100}   & 86.493 & 60.857 & 67.712 & 53.82    \\
\textbf{quantity-1000}  & 86.166 & 61.168 & 66.920 & 51.91    \\
\textbf{quantity-10000} & 86.753 & 61.550 & 66.713 & 52.29    \\ \hline
\end{tabular}
\caption{Experiment results on desirable properties of $S^{a(i)}$ as detailed in the section on Improving Prototypes. 
% We debias the PLM using \textbf{ADEPT} with different sets of $S^{a(i)}$ and compare the performance of each model on StereoSet-Intrasentence (from column 1 to 3) and CrowS-Pairs (column 4). \textbf{raw}, the original $S^{a(i)}$; \textbf{reliability}, modified $S^{a(i)}$, filtering sentences $S^{a(i)}_m$ if $len(S^{a(i)}_m)<30$; \textbf{quality}, sliced $S^{a(i)}$ ,so that $len(S^{a(1)}_m)=len(S^{a(2)}_m)=...$ for all pairwise attribute words; \textbf{quantity-$k$}, sliced $S^{a(i)}$, each with $k$ randomly sampled sentences.
}
\label{tbl:ab-tests}
\end{table}
We feed all sentences in $S^{a(i)}$ into the PLM, average the hidden states of the attribute words, and get a prototype $e^{a(i)}$ of attribute $a(i)$ on the manifold. As we aim to drive pairwise attribute words closer on the manifold, a prototype has to be clear and precise for generalizing the attribute's concept. Therefore, we perform several experiments adjusting the properties of $S^{a(i)}$ to improve $e^{a(i)}$. \textbf{raw} denotes the original $S^{a(i)}$. 

\subsubsection{Reliability} 
As contextualized word embeddings mix context information into every token's hidden states, for word $w^{a(i)}_m$, we need a myriad of context sentences to construct its prototype. Therefore, we regard $S^{a(i)}_m$ with $len(S^{a(i)}_m)<30$ as unreliable, and remove them from $S^{a(i)}$. 

\subsubsection{Quality}
Pairwise attribute words, like ``waiter" and ``waitress" for gender, should make equal contributions to the prototype. For a word $w^{a(i)}_m$, if its $S^{a(i)}_m$ makes up most of the $S^{a(i)}$, then the calculated prototype may well be influenced by the word's semantic meaning and leads to ambiguity. Thus, we enforce $len(S^{a(1)}_m)=len(S^{a(2)}_m)=...$ for all pairwise attribute words. 

\subsubsection{Quantity}
A larger corpus indicates more diverse training sentences and attribute words, but is more time-consuming to train. Hence, we test $S^{a(i)}$ with sizes at different orders of magnitude and compare the effects to choose the most desirable corpus size.

We run \textbf{ADEPT-finetuning} on the corpora mentioned above, stop the training at 500 steps (an early stage), and evaluate the debiased models on StereoSet-Intrasentence and CrowS-Pairs. Results are listed in Table \ref{tbl:ab-tests}. Data show that setting threshold for $S^{a(i)}_m$ and slicing pairwise $S^{a(i)}_m$ to be of equal size help improve the performance. In our experiments, we filter $S^{a(i)}_m$ if $len(S^{a(i)}_m)<30$, set $len(S^{a(1)}_m)=len(S^{a(2)}_m)=...$ and choose \textbf{quantity-10000}.

\section{Conclusion}
We proposed \textbf{ADEPT}, an algorithm that adopts prompt tuning for debiasing and introduces a new debiasing criterion inspired by manifold learning. By using prompt tuning, \textbf{ADEPT} consumes less computing and storage resources while preserving the base model's parameters, ensuring the model's robustness for other tasks after debiasing. Using this new debiasing criterion, \textbf{ADEPT} obtains competitive scores on bias benchmarks and even improves a PLM's representation ability for downstream tasks. By visualizing the words' correlation before and after the PLM is debiased, we find that \textbf{ADEPT} drives pairwise attributes closer on the manifold and keeps words' relative distances. \textbf{ADEPT} provides a smoother loss function than previous methods, allowing for better use of optimizations like early stopping. We also establish the standard of evaluating the corpus for building attribute prototypes in the contextualized word embedding setting and refine \textbf{ADEPT}'s performance with it. In the future, we will explore time-efficient objective terms for keeping words' relative distances in debiasing and release a new dataset that measures a model's biases and expressiveness comprehensively, free of the need to make predictions about masked tokens.

\section{Ethics Statement}
When designing \textbf{ADEPT}, we made some assumptions to simplify the complex world model which may lead to some ethical concerns. 
% Considering the ethical scope of our work, we will discuss the validity and potential implications of these assumptions.

% We only tested \textbf{ADEPT} on language models trained for English and 
We defined bias in this paper as the difference between attribute prototypes relative to neutral words. However, this requires carefully selecting the categories of the attribute based on real-world debiasing demands. Unfortunately, our construction could be reversed such that the words in the attributes list are made as different in distance from the neutral words as possible, but we expect that this would cause the embeddings to degrade and thus not be effective for intentionally causing harm. 

In our paper, we discussed the usage of \textbf{ADEPT} on the binary gender setting, which in general is not reflective of the real world, where gender (and other biases) can be far from binary. It is reasonable to have concerns that a binary construction can cause harms to groups not part of the pair. Luckily, all pieces of \textbf{ADEPT} are directly extensible to any number of dimensions, allowing for all dimensions to be pushed to cluster together. 

% couldn't suffice to cover all biases under this topic. Moreover, we aggregate attribute prototypes from manually selected pairwise words, where the hand-crafted attribute word pairs would inevitably introduce subjectivity to the debiasing process. Additionally, this work doesn't explore the intersectionality of stereotyping, such as how bias takes form if we take both social class and gender into account.
Unfortunately, we cannot ensure or contradict causality between bias reduction and discrimination mitigation in PLMs. \citet{goldfarb2020intrinsic} makes an effort to deny the causality between bias and discrimination, but it only takes WEAT as the intrinsic task, so more work is needed in this area to ensure that we are indeed reducing harms. 
% The debiasing space is expecting theoretical guidance and experimental proof in this direction.

% Fourth, even \textbf{ADEPT} fails to reduce all the biases measured by the benchmarks we have included. More importantly, further tuning the model may reproduce, or even amplify the pre-defined bias. Therefore, \textbf{ADEPT} is never a one-size-fits-all approach. After debiasing with \textbf{ADEPT}, researchers should critically investigate the discrimination risk when implementing the debiased model in real-world tasks.

\section{Acknowledgements}
We thank the anonymous reviewers' helpful suggestions.
% This research is based upon work supported in part by U.S. DARPA AIDA Program No. FA8750-18-2-0014 and U.S. DARPA KAIROS Program Nos. FA8750-19-2-1004. 
% The views and conclusions contained herein are those of the authors and should not be interpreted as necessarily representing the official policies, either expressed or implied, of DARPA, or the U.S. Government. The U.S. Government is authorized to reproduce and distribute reprints for governmental purposes notwithstanding any copyright annotation therein.

\section{Reference}
\label{sec:reference}
\nobibliography*
\bibentry{9799257}.\\[.2em]
\bibentry{askell2021general}.\\[.2em]
\bibentry{bentivogli2009fifth}.\\[.2em]
\bibentry{blodgett2020language}. \\[.2em]
\bibentry{NIPS2016_a486cd07}.\\[.2em]
\bibentry{bommasani-etal-2020-interpreting}.\\[.2em]
\bibentry{bookcorpus_hf}.\\[.2em]
\bibentry{caliskan2017semantics}.\\[.2em]
\bibentry{dagan2005pascal}.\\[.2em]
\bibentry{dai2015semi}.\\[.2em]
\bibentry{devlin2018bert}.\\[.2em]
\bibentry{dolan-brockett-2005-automatically}.\\[.2em]
\bibentry{giampiccolo2007third}.\\[.2em]
\bibentry{goldfarb2020intrinsic}.\\[.2em]
\bibentry{guo-etal-2022-auto}.\\[.2em]
\bibentry{haim2006second}.\\[.2em]
\bibentry{hinton2002stochastic}.\\[.2em]
\bibentry{kaneko-bollegala-2019-gender}.\\[.2em]
\bibentry{kaneko2021debiasing}.\\[.2em]
\bibentry{kingma2014adam}.\\[.2em]
\bibentry{lester2021power}.\\[.2em]
\bibentry{levesque2012winograd}.\\[.2em]
\bibentry{li2021prefix}.\\[.2em]
\bibentry{liang2020towards}.\\[.2em]
\bibentry{liu2021gpt}.\\[.2em]
\bibentry{liu2021p}.\\[.2em]
\bibentry{manzini2019black}.\\[.2em]
\bibentry{may2019measuring}.\\[.2em]
\bibentry{mayfield2019equity}.\\[.2em]
\bibentry{meade2021empirical}. \\[.2em]
\bibentry{melas2020mathematical}.\\[.2em]
\bibentry{mikolov2013efficient}.\\[.2em]
\bibentry{mikolov2013linguistic}.\\[.2em]
\bibentry{nadeem2020stereoset}.\\[.2em]
\bibentry{nangia2020crows}.\\[.2em]
\bibentry{10.2307/2237880}.\\[.2em]
\bibentry{peters-etal-2018-deep}.\\[.2em]
\bibentry{rabelo2022overview}.\\[.2em]
\bibentry{radford2018improving}.\\[.2em]
\bibentry{raffel2020exploring}.\\[.2em]
\bibentry{schick2021self}.\\[.2em]
\bibentry{socher-etal-2013-recursive}.\\[.2em]
\bibentry{solaiman2021process}.\\[.2em]
\bibentry{TIEDEMANN12.463}.\\[.2em]
\bibentry{wang2018glue}.\\[.2em]
\bibentry{wolf2019huggingface}.\\[.2em]
\bibentry{yang2019xlnet}.\\[.2em]
\bibentry{zhao2018learning}.\\[.2em]
\bibentry{Zhu_2015_ICCV}.\\[.2em]
\bibentry{zmigrod2019counterfactual}.
\nobibliography{aaai23}
% \clearpage
% \section{Appendix} \label{appendix}
% \begin{figure}[h]
%     \centering
%     \subfloat[\textbf{original}]{\includegraphics[width=1.62in]{Figure3-a.png} \label{3-a}}
%     \hfill
%     \subfloat[\textbf{ADEPT}]{\includegraphics[width=1.62in]{Figure3-b.png} \label{3-b}}
%     \caption{Visualized correlation of words in the religion domain. We use t-SNE to plot the figures and set perplexity as 80. We color neutral words grey, Judaism words yellow, Christianity words blue, and Islam words green.}
%     \label{fig:3}
% \end{figure}
% \begin{figure}[h]
%     \centering
%     \subfloat[\textbf{DPCE}]{\includegraphics[width=1.62in]{Figure4-a.png} \label{4-a}}
%     \hfill
%     \subfloat[\textbf{ADEPT}]{\includegraphics[width=1.62in]{Figure4-b.png} \label{4-b}}
%     \caption{Evaluation loss in the training processes for \textbf{DPCE} and \textbf{ADEPT}.}
%     \label{fig:4}
% \end{figure}
\end{document}